\documentclass[sigconf, authorversion=true]{acmart}
\usepackage{subcaption}

\AtBeginDocument{%
  \providecommand\BibTeX{{%
    \normalfont B\kern-0.5em{\scshape i\kern-0.25em b}\kern-0.8em\TeX}}}

\copyrightyear{2022}
\acmYear{2022}
\setcopyright{acmcopyright}

\acmConference[HAI '22] {Proceedings of the 10th International Conference on Human-Agent Interaction }{December 5--8, 2022}{Christchurch, New Zealand}

\acmBooktitle{Proceedings of the 10th International Conference on Human-Agent Interaction (HAI '22), December 5--8, 2022, Christchurch, New Zealand}

\acmPrice{15.00}

\acmDOI{10.1145/3527188.3563915}
\acmISBN{978-1-4503-9323-2/22/12}

\begin{document}

\title{Interaction in Remote Peddling Using Avatar Robot by People with Disabilities}

\author{Takashi Kanetsuna}
\email{s1912021@cco.kanagawa-it.ac.jp}
\affiliation{%
  \institution{Kanagawa Institute of Technology}
  \city{Atsugi}
  \state{Kanagawa}
  \country{Japan}
}

\author{Kazuaki Takeuchi}
\email{k.takeuchi@orylab.com}
\affiliation{%
  \institution{OryLab Inc.}
  \city{Chuo-ku}
  \state{Tokyo}
  \country{Japan}
}

\author{Hiroaki Kato}
\email{h.kato@orylab.com}
\affiliation{%
  \institution{OryLab Inc.}
  \city{Chuo-ku}
  \state{Tokyo}
  \country{Japan}
}

\author{Taichi Sono}
\email{taichisono0420@keio.jp}
\orcid{}
\affiliation{%
  \institution{Keio University}
  \city{Yokohama}
  \state{Kanagawa}
  \country{Japan}
}

\author{Hirotaka Osawa}
\email{osawa@ae.keio.ac.jp}
\orcid{0000-0001-5779-8437}
\affiliation{%
  \institution{Keio University}
  \city{Yokohama}
  \state{Kanagawa}
  \country{Japan}
}

\author{Kentaro Yoshifuji}
\affiliation{%
  \institution{OryLab Inc.}
  \city{Chuo-ku}
  \state{Tokyo}
  \country{Japan}
}
\author{Yoichi Yamazaki}
\email{yamazaki@he.kanagawa-it.ac.jp}
\orcid{0000-0001-6402-4587}
\affiliation{%
  \institution{Kanagawa Institute of Technology}
  \city{Atsugi}
  \state{Kanagawa}
  \country{Japan}
}

\renewcommand{\shortauthors}{Kanetsuna and Takeuchi, et al.}

\begin{abstract}
  Telework "avatar work," in which people with disabilities can engage in physical work such as customer service, is being implemented in society. In order to enable avatar work in a variety of occupations, we propose a mobile sales system using a mobile frozen drink machine and an avatar robot “OriHime”, focusing on mobile customer service like peddling. The effect of the peddling by the system on the customers are examined based on the results of video annotation.
\end{abstract}



\begin{teaserfigure}
 \begin{minipage}{0.31\linewidth}
  \centering
  \includegraphics[height=33mm]{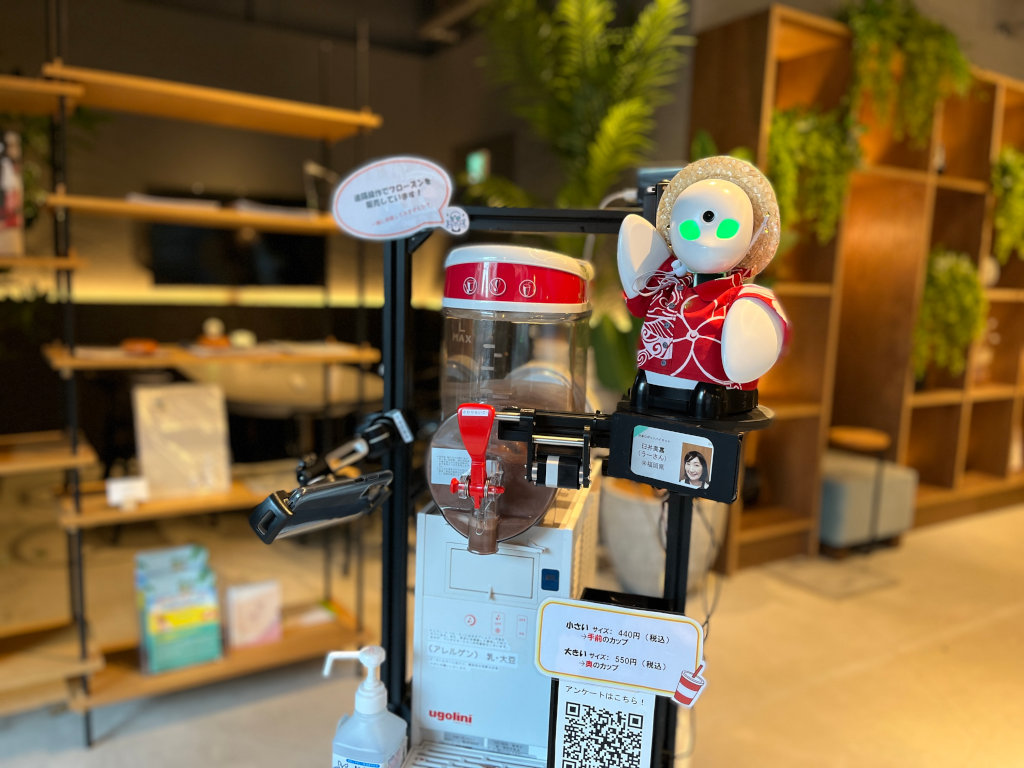}
  \subcaption{mobile frozen machine with OriHime}
  \label{fig:one}
 \end{minipage}
 \begin{minipage}{0.32\linewidth}
  \centering
  \includegraphics[height=33mm]{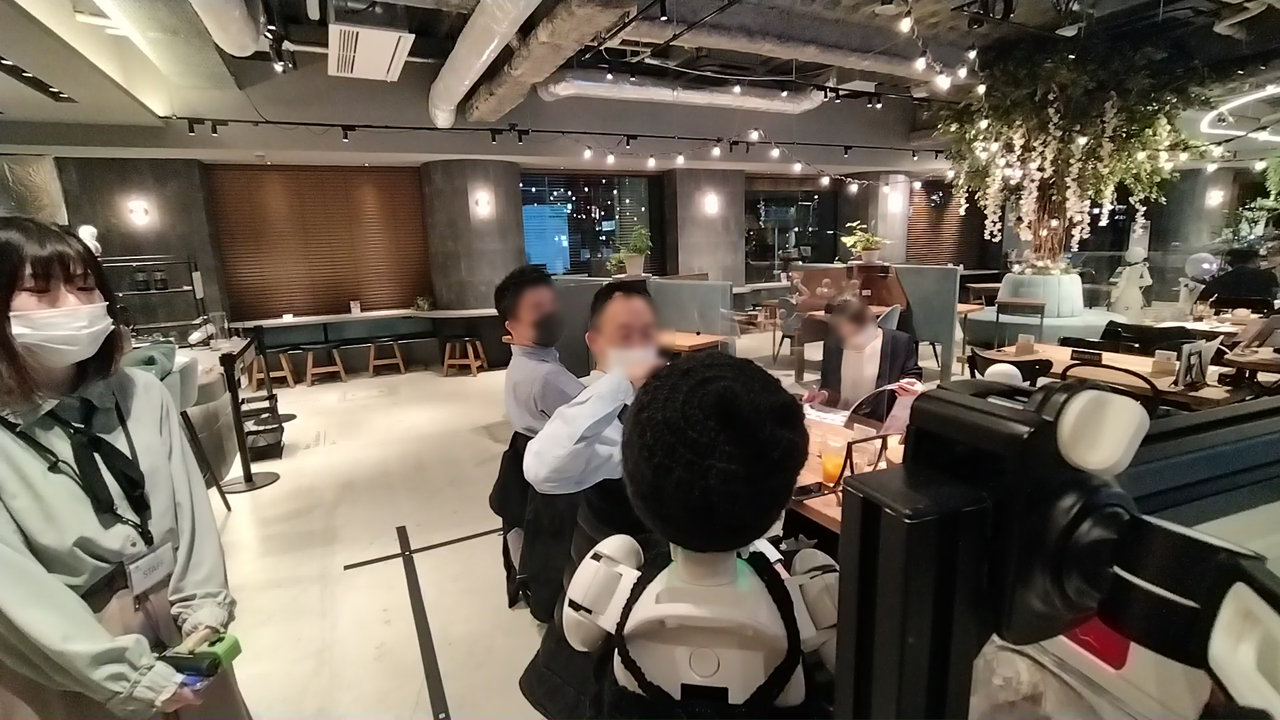}
  \subcaption{experimental scene (cafe side)}
  \label{fig:two}
 \end{minipage}
 \begin{minipage}{0.36\linewidth}
  \centering
  \includegraphics[height=33mm]{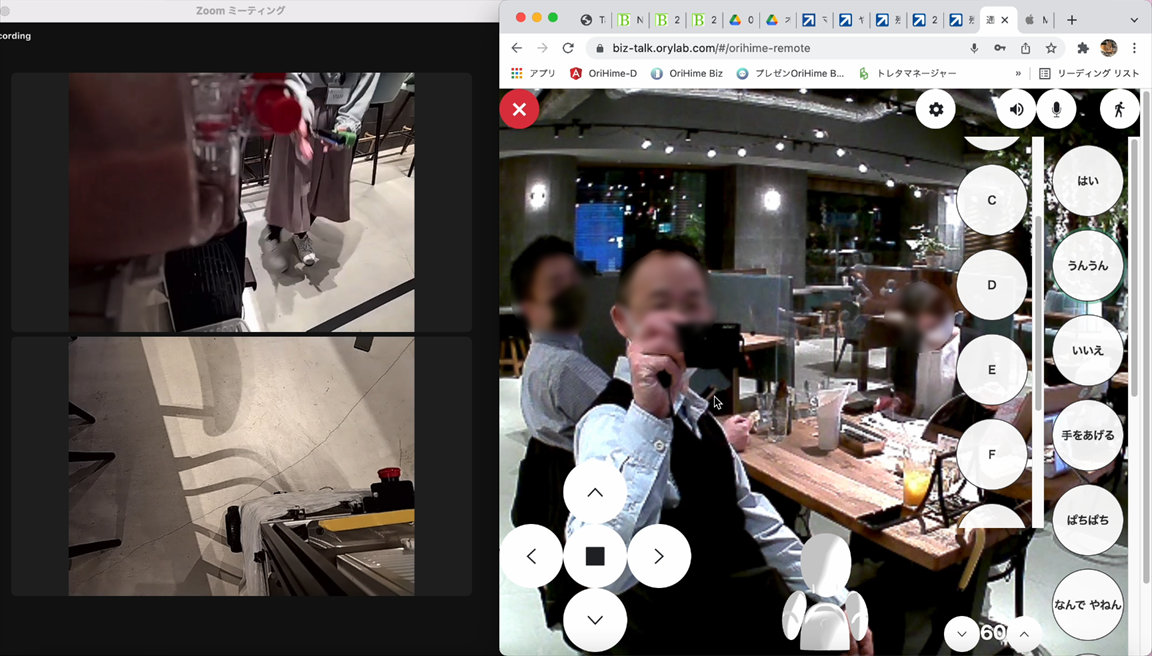}
  \subcaption{experimental scene (pilot side)}
  \label{fig:three}
 \end{minipage}
 \caption{Operation screens by a pilot who has disabilities at the avatar robot cafe DAWN ver. $\beta$.}
 \Description{Customer service scenes at Avatar robot cafe. The pilots are serving customers with physical and virtual cybernetic avatars}
 \label{fig:teaser}
\end{teaserfigure}

\received{3 August 2022}
\received[revised]{25 September 2022}
\received[accepted]{10 September 2022}

\maketitle

\section{Introduction}
As of 2022, the legally mandated employment rate for people with physical disabilities in Japan is set at 2.30\%. On the other hand, the actual employment rate is 2.20\%, which means that more employment opportunities need to be created \cite{MHLW21}. We focused on remote employment using telepresence robots as a means of creating employment opportunities for people with disabilities. As remote employment by avatar robots, "avatar work" has been realized in which people with disabilities use OriHime-D to engage in customer service and physical work \cite{Takeuchi20}. In this context, passive customer service is implemented by responding to requests for cafe visitors and serving them food. On the other hand, mobile proposal-based customer service, in which an avatar robot actively  engages and sells while moving around like a peddler, has not yet been realized.

To realize an inclusive society through the avatar work of diverse professions, we propose a mobile sales system using a mobile frozen drink machine and OriHime as a mobile proposal-type customer service system. The influence of the operator on the customers for the mobile sales will be examined based on the results of video annotation.

\section{Avatar work by people with disabilities}
Avatar robots have been developed to enable social participation for people with disabilities who have difficulty going out. Focusing on employment as a form of social participation, Avatar Work, a remote employment using an avatar robot, and an avatar robot cafe have been proposed as a place for this purpose and are already operating in actual stores \cite{Takeuchi20}\cite{DAWN21}. In the avatar robot cafe, the operator with disabilities (pilot) extends his/her own physical and spatial abilities using two types of avatar robots, a stationary type and a mobile type, and is able to respond to customer orders and serve food to them. Figure 2 shows the stationary type avatar robot OriHime and the mobile abatar robot OriHime-D \cite{OriHime}. These robots are equipped with a camera and microphone, and the operator can communicate with a person in front of the robot via the Internet. These alter ego robots are equipped with cameras and microphones, and the pilot can communicate with a person in front of the robot via the Internet.

In this study, we focused on proposal-based customer service, in which the pilot proactively proposes products and services. If proposal-based customer service that involves physical action becomes possible, avatar work in a variety of occupations can be realized, increasing both employment opportunities and options for people with disabilities. The ability to serve customers with proposals that involve physical action, such as peddling, would allow for a diverse range of jobs to be avatarworked, increasing both employment opportunities and options for people with disabilities. Creating an environment where people with disabilities can work comfortably will lead to a society that is inclusive of people with disabilities.
For proposal-based customer service with action, it is necessary to realize two functions: communication and sales behavior including movement. For this purpose, we developed a peddling robot system that combines the avatar robot OriHime with a sales function, and prepared an operating interface for the system.

\begin{figure}[h]
 \begin{minipage}{0.6\linewidth}
  \centering
  \includegraphics[height=33mm]{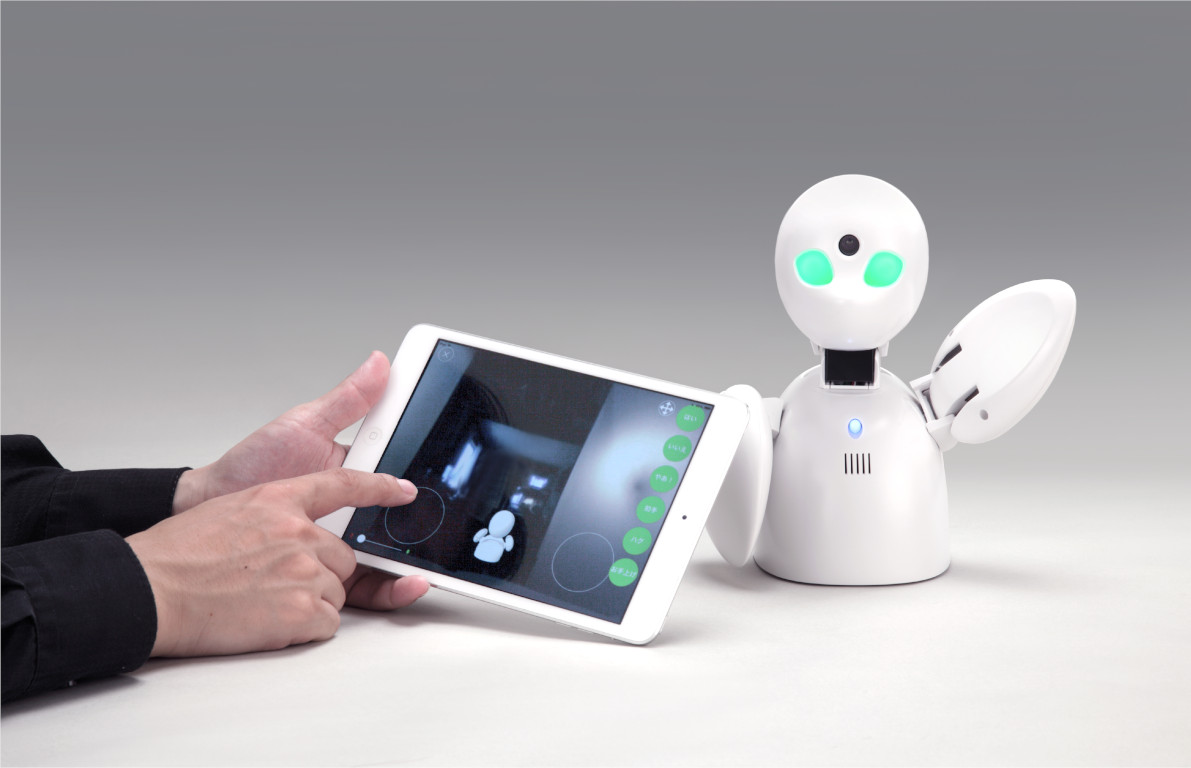}
  \subcaption{OriHime}
  \label{fig:one}
 \end{minipage}
 \begin{minipage}{0.35\linewidth}
  \centering
  \includegraphics[height=33mm]{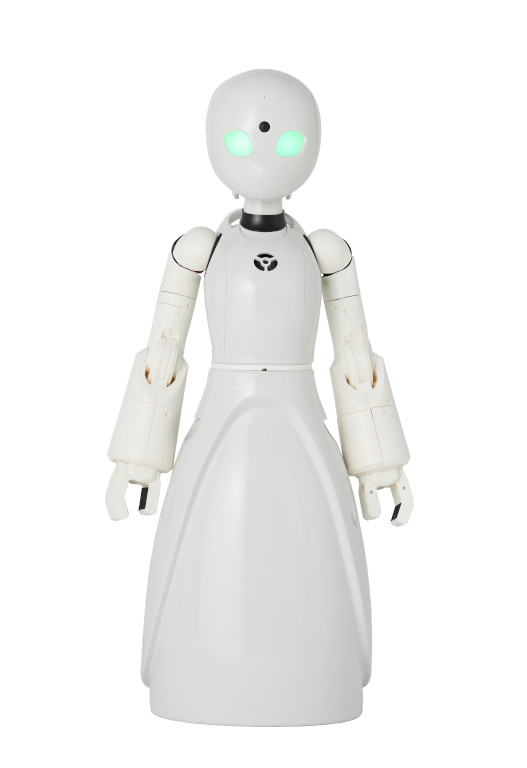}
  \subcaption{OriHime-D}
  \label{fig:two}
 \end{minipage}
 \caption{OriHime and OriHime-D.}
\end{figure}

\section{Peddling using avatar robot}

\begin{figure}[h]
  \centering
  \includegraphics[width=\linewidth]{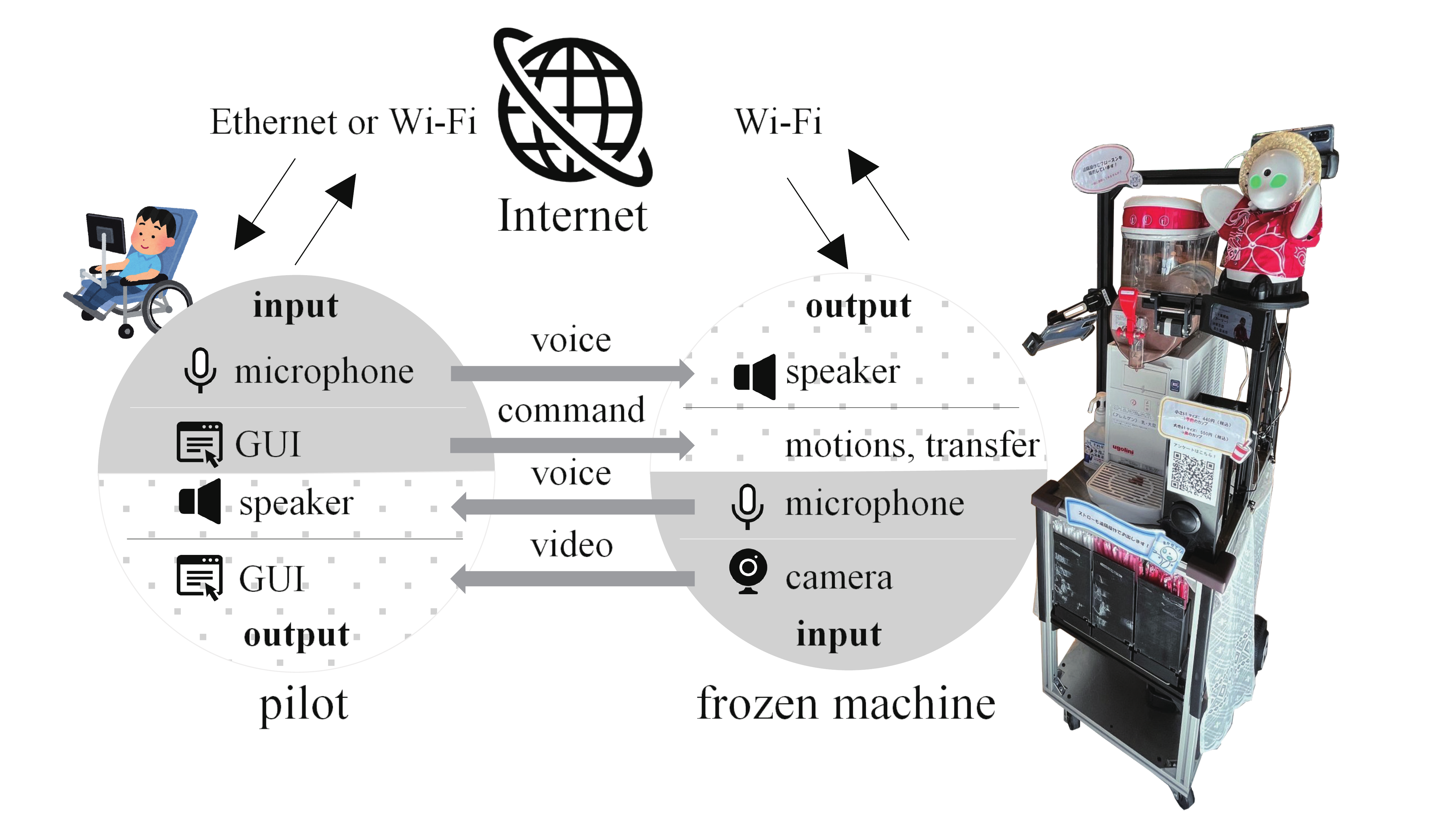}
  \caption{Operation of the avatar robot OriHime-D by a pilot with disabilities at the Avatar robot cafe.}
  \Description{A pilot with disability operates OriHime-D at the cafe, from her home.}
\end{figure}

As a proposal-based customer service in which the pilot proactively proposes products and services using the avatar robot, we prepared a mobile frozen drink vending machine inside the avatar robot cafe. The mobile frozen drink machine developed for the peddling is shown in Figure 1 (a). The pilot drives the mobile frozen machine and communicates with customers via OriHime installed in the machine. An overview of the operation flow of the proposed system is shown in Figure 3. The combination of a mobile frozen machine and OriHime enables cooperative actions while communicating, such as pouring a drink into a cup supported by a customer.

The pilot operates the mobile frozen machine with OriHime installed in the Avatar Robot Cafe from home via the Internet. To enable operation for cooperative behavior while communicating, we prepared an interface with multiple viewpoints: a viewpoint for the communication function and a viewpoint for the action.  Figure 1 (c) shows the pilot's operation screen and Figure 1 (b) shows the scene at the cafe.

\section{Evaluation of mobile sales-type avatar work}
A mobile sales-type avatar work with the proposed system was conducted to evaluate the impact on the interaction between the pilot and the customers.

\subsection{Experimental Procedure}
A mobile sales-type avatar work with the proposed system was conducted to evaluate the impact on the interaction between the pilot and the customers.

In the experiment, the pilot operated a mobile frozen machine and moved to each table in the experimental space (four seats per table, eight tables with 32 seats in total) in the avatar robot cafe, approached customers who were eating, and asked them to buy a frozen drink. When a customer purchases a drink, the customer first selects a cup, following the pilot's instructions. Next, the pilot and customer work together to pour the drink into the cup. Finally, the pilot opens the straw box provided and asks the customer to select a straw. The payment is collected by the assistant from the customer. The whole process, both on the cafe side and on the pilot's side, were captured on video. A questionnaire was also administered to customers who purchased the drinks. The evaluation was based on the questionnaire and annotation analysis of the filmed videos. The experiment was conducted at an avatar robot cafe in Nihonbashi, Tokyo, for five pilots from March 15 to April 1, 2022 (9 working days), with working hours ranging from 1 to 3 hours per day. The study was conducted for five pilots at the robot cafe. For the experiment, an assistant was provided to assist with the operation of the robot. The assistant assists the robot's movement and communication as needed. The assistant was not involved in communicating with customers or serving frozen food, but was involved in restoring mobility and communication.In annotating the videos taken, we focused on the number of times 16 different items were executed and the elapsed time.

\subsection{Results}
This paper presents the results of the annotation analysis, focusing on the work of one pilot (4 working days, 9 hours in total). The pilot had coronary angina and was piloting from home.

First, focusing on the amount of interaction (frequency and time) between the customers and the avatar robot, it was observed that customers who shared more eye contact with their companions also shared more eye contact with OriHime. This suggests that people with high social skills treat OriHime as if it were a person.

In addition, as a result of interviews via questionnaires, customers commented that they enjoyed working together with the robot. The collaborative process of pouring drinks encourages physical communication and may lead to enjoyment, even in cases where it did not go well. Given the positive comments received from customers, we have determined that the proposed system can be operated continuously when the avatar robot peddles, and we plan to continue the experiment in the future.

The next issue is that the role of assistants is essential for movement and communication. Manual operation in a real environment places a heavy operational burden on the pilot. In the next step, the role of the assistant should be analyzed, functionalized as needed, and incorporated into the system for semi-automation.

\section{CONCLUSIONS}

To realize an inclusive society through the avatar work of diverse professions, we proposed a mobile vending service using a mobile frozen drink machine and OriHime as a mobile proposal type customer service. As a result of the verification, it was confirmed that peddling by OriHime was received in the same way as a person and that the collaborative work was positively received. 

Currently, "AVATAR GUILD" has been established as a work support service for people with disabilities that allows them to work remotely using an avatar robot, and the number of entries by people with disabilities is increasing \cite{AVATARGUILD}. The proposed system will lead to the diversification of job types that can be introduced into the AVATAR GUILD.

\begin{acks}
  This work was supported by JST Moonshot R\&D Program “Cybernetic being” Project (Grant number JPMJMS2013).
\end{acks}

\bibliographystyle{ACM-Reference-Format}
\bibliography{HAI-kanetsuna}


\begin{thebibliography}{5}


\ifx \showCODEN    \undefined \def \showCODEN     #1{\unskip}     \fi
\ifx \showDOI      \undefined \def \showDOI       #1{#1}\fi
\ifx \showISBNx    \undefined \def \showISBNx     #1{\unskip}     \fi
\ifx \showISBNxiii \undefined \def \showISBNxiii  #1{\unskip}     \fi
\ifx \showISSN     \undefined \def \showISSN      #1{\unskip}     \fi
\ifx \showLCCN     \undefined \def \showLCCN      #1{\unskip}     \fi
\ifx \shownote     \undefined \def \shownote      #1{#1}          \fi
\ifx \showarticletitle \undefined \def \showarticletitle #1{#1}   \fi
\ifx \showURL      \undefined \def \showURL       {\relax}        \fi
\providecommand\bibfield[2]{#2}
\providecommand\bibinfo[2]{#2}
\providecommand\natexlab[1]{#1}
\providecommand\showeprint[2][]{arXiv:#2}

\bibitem[MHLW(2021)]%
        {MHLW21}
\bibfield{author}{\bibinfo{person}{MHLW}.} \bibinfo{year}{2021}\natexlab{}.
\newblock \bibinfo{title}{Aggregated Data on the Employment Status of Persons
  with Disabilities (in Japanese)}.
\newblock
\newblock
\urldef\tempurl%
\url{https://www.mhlw.go.jp/content/11704000/000871748.pdf}
\showURL{%
\tempurl}
\newblock
\shownote{\url{https://www.mhlw.go.jp/content/11704000/000871748.pdf}}.


\bibitem[OryLab.(2010)]%
        {OriHime}
\bibfield{author}{\bibinfo{person}{OryLab.}} \bibinfo{year}{2010}\natexlab{}.
\newblock \bibinfo{title}{Avatar robot OriHime}.
\newblock
\newblock
\urldef\tempurl%
\url{https://orylab.com/en/}
\showURL{%
\tempurl}
\newblock
\shownote{\url{https://orylab.com/en/}}.


\bibitem[OryLab.(2021a)]%
        {AVATARGUILD}
\bibfield{author}{\bibinfo{person}{OryLab.}} \bibinfo{year}{2021}\natexlab{a}.
\newblock \bibinfo{title}{AVATAR GUILD}.
\newblock
\newblock
\urldef\tempurl%
\url{https://avatarguild.com/}
\showURL{%
\tempurl}
\newblock
\shownote{\url{https://avatarguild.com/}}.


\bibitem[OryLab.(2021b)]%
        {DAWN21}
\bibfield{author}{\bibinfo{person}{OryLab.}} \bibinfo{year}{2021}\natexlab{b}.
\newblock \bibinfo{title}{Avatar robot cafe DAWN ver. $\beta$}.
\newblock
\newblock
\urldef\tempurl%
\url{https://dawn2021.orylab.com/en/}
\showURL{%
\tempurl}
\newblock
\shownote{\url{https://dawn2021.orylab.com/en/}}.


\bibitem[Takeuchi et~al\mbox{.}(2020)]%
        {Takeuchi20}
\bibfield{author}{\bibinfo{person}{Kazuaki Takeuchi}, \bibinfo{person}{Yoichi
  Yamazaki}, {and} \bibinfo{person}{Kentaro Yoshifuji}.}
  \bibinfo{year}{2020}\natexlab{}.
\newblock \showarticletitle{Avatar Work: Telework for Disabled People Unable to
  Go Outside by Using Avatar Robots ``OriHime-D'' and Its Verification}. In
  \bibinfo{booktitle}{\emph{Companion of the 2020 ACM/IEEE International
  Conference on Human-Robot Interaction}} \emph{(\bibinfo{series}{HRI '20})}.
  \bibinfo{publisher}{Association for Computing Machinery},
  \bibinfo{address}{New York, NY, USA}, \bibinfo{pages}{53–60}.
\newblock
\urldef\tempurl%
\url{https://doi.org/10.1145/3371382.3380737}
\showDOI{\tempurl}


\end{thebibliography}

\end{document}